\documentclass[conference]{IEEEtran}

\usepackage{amsmath} 
\usepackage{amssymb}  
\usepackage{dsfont}
\usepackage{graphicx}

\usepackage{psfrag,graphicx,epsfig}
\usepackage{epstopdf}
\usepackage{xspace}
\usepackage{subfig}
\usepackage{float}
\usepackage{placeins}
\usepackage{multirow}
\usepackage{pgf,tikz}
\usepackage{nowidow}
\usepackage{lineno}
\usepackage{xcolor}
\usepackage{arydshln} 

\usepackage{siunitx}
\usepackage{color}
\usepackage{flushend}
\usepackage{lineno}
\usepackage{algorithmic}

\usepackage[vlined,linesnumbered,boxruled]{algorithm2e}

\usepackage{tikz} 
\usepackage{tikz-3dplot}

\usepackage[numbers]{natbib}
\usepackage{multicol}
\usepackage[bookmarks=true]{hyperref}

\def\Gamma{20}

\usepackage{caption}
\usepackage{booktabs}
\usepackage{varwidth}

\usepackage{wrapfig}

\usepackage{subfig}

\begin{document}

\title{\LARGE \bf
Understanding Multi-Modal Perception Using Behavioral Cloning for Peg-In-a-Hole Insertion Tasks 
}


\author{\authorblockN{Yifang Liu}
\authorblockA{Electrical and Computer\\ Engineering,
Cornell University\\
Ithaca, NY 14853\\
Email: yl892@cornell.edu}
\and
\authorblockN{Diego Romeres}
\authorblockA{Mitsubishi Electric\\ Research Laboratories\\
Cambridge, MA 02139\\
Email: romeres@merl.com}
\and
\authorblockN{Devesh K. Jha}
\authorblockA{Mitsubishi Electric\\ Research Laboratories\\
Cambridge, MA 02139\\
Email: jha@merl.com}
\and
\authorblockN{Daniel Nikovski}
\authorblockA{Mitsubishi Electric\\ Research Laboratories\\
Cambridge, MA 02139\\
Email: nikovski@merl.com}}


\maketitle

\begin{abstract}
One of the main challenges in peg-in-a-hole (PiH) insertion tasks is in handling the uncertainty in the location of the target hole. In order to address it, high-dimensional sensor inputs from sensor modalities such as vision, force/torque sensing, and proprioception can be combined to learn control policies that are robust to this uncertainty in the target pose. 
Whereas deep learning has shown success in recognizing objects and making decisions with high-dimensional inputs, 
the learning procedure might damage the robot when applying directly trial-and-error algorithms on the real system.
At the same time, learning from Demonstration (LfD) methods have been shown to achieve compelling performance in real robotic systems by leveraging demonstration data provided by experts.
In this paper, we investigate the merits of multiple sensor modalities such as vision, force/torque sensors, and proprioception when combined to learn a controller for real world assembly operation tasks using LfD techniques. The study is limited to PiH insertions; we plan to extend the study to more experiments in the future. Additionally, we propose a multi-step-ahead loss function to improve the performance of the behavioral cloning method.
Experimental results on a real manipulator support our findings, and show the effectiveness of the proposed loss function. 
\end{abstract}
%
%
%
\section{Introduction}
\label{sec:Introduction}

Contact-rich manipulation tasks, such as peg-in-a-hole (PiH) insertion, are very common in our daily lives, and can be performed effortlessly by humans. 
In contrast, it is still difficult for robots to perform these simple tasks when the target location is not specified accurately. 
If the target location is known exactly, methods such as motion planning and Dynamic Movement Primitives (DMP) \cite{00393,5152385} can perform these tasks well, in combination with standard compliance controllers. However, these methods would typically fail if non-negligible uncertainty is present in the target location. 
Recently, vision-guided manipulation methods based on deep learning or deep reinforcement learning have received much attention, because they are robust to the variation of the target location \cite{lee2019making,DBLP:journals/corr/abs-1906-05841,vecerik2019practical}.
However, these algorithms often require large amounts of training samples, good exploration strategies, or a carefully designed loss function due to the high dimensionality and non-linearity of the sensor inputs.

\begin{figure}[t]
\centering
\includegraphics[scale=0.32]{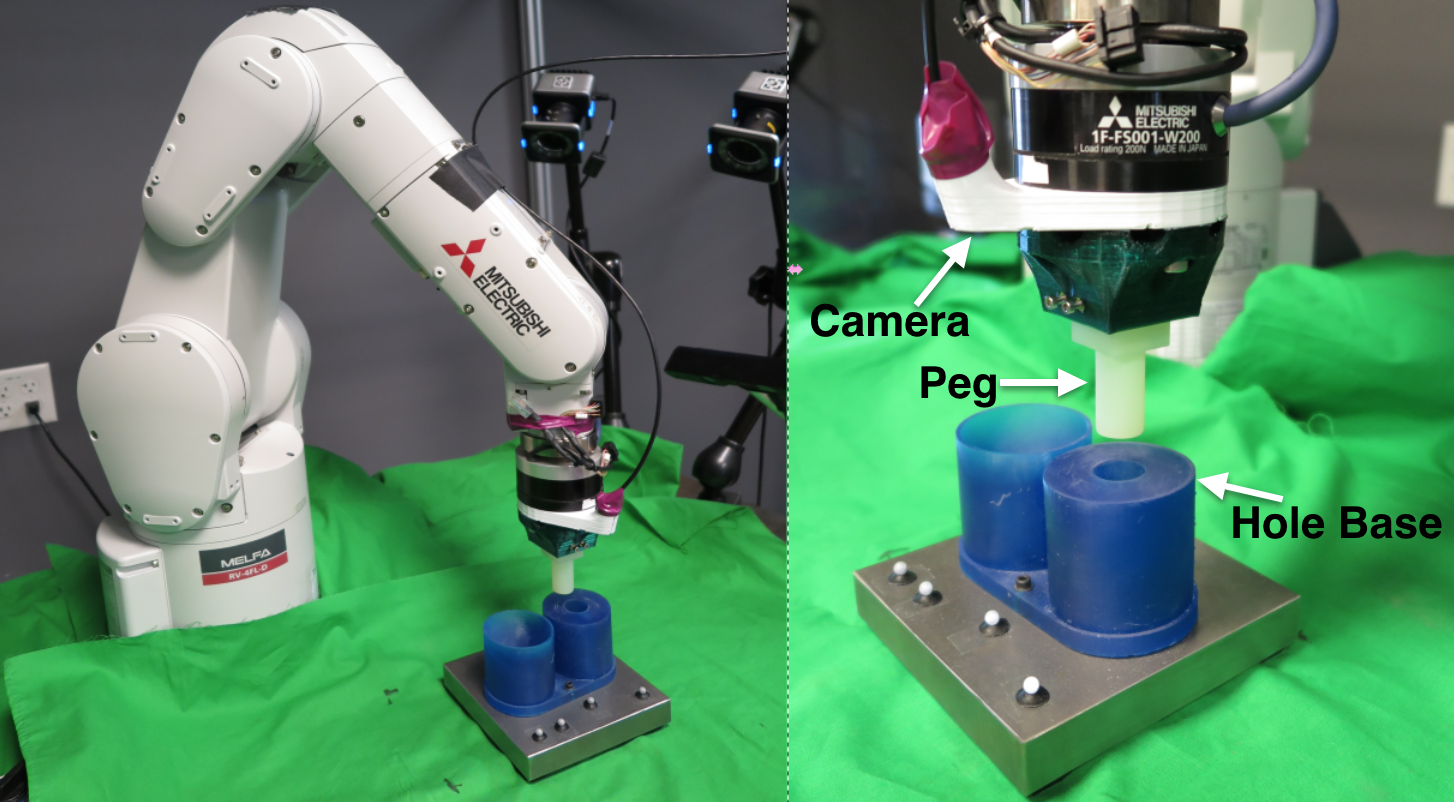}
\caption{Peg-in-a-hole experimental setup. Left: a Mitsubishi Electric RV-4FL model robot arm, and a 3D-printed cylindrical peg, a green peg holder, a white wrist camera holder (attached on the end-effector), and a blue hole base. Right: A closer view of the peg, the hole base, and the wrist camera (located underneath the purple tape with black wire).} 
\label{fig:setup}
\end{figure}

Learning from Demonstration (LfD) is a useful tool to speed up training, as it allows robots to leverage expert demonstration data to efficiently learn skills to perform manipulation tasks with acceptable performance. It can also be used to warm-start reinforcement learning techniques \cite{DBLP:journals/corr/HesterVPLSPSDOA17}. LfD has been applied successfully to a wide range
of domains in robotics, for example, in autonomous driving \cite{DBLP:journals/corr/BojarskiTDFFGJM16,7358076}
and manipulation \cite{akgunkeyframe}.
Most LfD methods use humans as demonstrators \cite{argall2009survey}. However, human demonstrations are usually noisy and time-consuming. 
For tasks such as PiH, if the target location is known, which is usually the case at training time, data collection can become easier by implementing a basic controller on the robot instead of using human demonstrations. For these reasons, in this work, we use an automated controller to collect demonstrations. Behavioral cloning is an efficient imitation learning method, which learns the policy used by the expert during demonstration given the state of the environment \cite{DBLP:journals/corr/abs-1905-11108,ijcai2018-687}. However, simple cloning models that are only trained on an expert's policy distribution are prone to mistakes due to the state distribution shift at test time, whereas more complex behavioral cloning methods that generalize to out-of-distribution states can be difficult to use \cite{DBLP:journals/corr/abs-1905-11108}. 
We propose an efficient and simple behavioral cloning approach for PiH, based on a multi-step-ahead loss function. Whereas multi-step-ahead loss functions, as opposed to the more common one-step-ahead loss functions, have been already proposed for model learning \cite{doerr2017optimizing,venkatraman2015improving}, to the best of our knowledge, they have never been proposed for model-free behavioral cloning algorithms. Their advantage is that, at each time step, the policies trained with these loss functions enforce the system to end up in states that are still close to the demonstrations.

The main effort of this work is in combining and experimenting with different sensor modalities when performing insertions task in robotic systems.
Indeed, various sensors such as cameras, force/torque (F/T) sensors, and positional sensors can (and should) be used both to provide feedback measurements as well as to give information about the target location when unknown. For instance, in PiH insertions, vision can provide information about the relative location of the hole w.r.t. the peg, and force-torque readings can monitor the forces exerted upon the peg allowing impedance and compliant control. Physics-based simulations, widely used in simulating body dynamics and collision detection, could also be a valid alternative to study the merits of different sensor modalities. However, the contact dynamics and friction properties during peg insertions are hard to model precisely in simulation. For this reason, understanding multi-modal perception on real robotic systems is still an open problem.

In practice, PiH tasks can be divided into two sub-tasks. The first is approaching the neighborhood of the hole with the peg, based on (possibly uncertain) information of the target location, and the second is the actual insertion into the hole.  
The former sub-task can be accomplished with standard trajectory planning techniques, or machine learning techniques such as DMP, when a demonstration is available. The approximate position of the hole can be known a priori or estimated by using motion-capture systems or off-the-shelf pose estimation techniques. As a result, the first sub-task can be solved relatively easily. 
The second sub-task is a more complex manipulation problem, where high precision control is required, and where predefined trajectories or open-loop controllers would most likely fail. 
Indeed, the controller needs to handle possible unforeseen impacts and contact forces. Understanding the role of different sensor modalities in real robotic systems is fundamental to successfully perform assembly tasks with significant uncertainty.

 For these reasons, we are interested in solving the second sub-task. Focusing on a sub-task has the advantage of requiring less data, and avoids the use of computational power and learning capability to learn components such as the approach trajectory, that can be more efficiently computed in other ways.

\subsection{Contributions}
Motivated by the above discussion, we investigate methods for learning different controllers for a robotic manipulator to perform peg insertions starting from a neighborhood of the target hole. We rely on a behavioral cloning approach comparing deep-learning-based controllers designed to handle different input modalities. We focus on empirically analyzing the importance of different sensor modalities, i.e., images from a wrist camera, end-effector
state information (proprioception), and force/torque measurements at the wrist of the robot. The first person point-of-view (p.o.v.) offers higher precision and every-day advantages such as simpler calibration procedures w.r.t. a third p.o.v. camera. Its use is also motivated by the recent increase in the number of commercial robots equipped with cameras at the wrist. The main open questions we investigate are: 1) Is it always advantageous to consider all three sensor modalities? 2) Can we achieve success even if we do not specify the target location of the hole in spatial coordinates?\\
%
%
The main contributions of the paper can be summarized as:
\begin{itemize}
 \item Empirical analysis of the effects of multi-modal perception in PiH insertions by combining several different sensor modalities into the control policy.
\item Introduction of a multi-step-ahead loss function to improve the performance of the policy learned from demonstration data in a behavioral cloning approach.
\item We obtained deep learning-based controllers that successfully perform PiH insertions and are goal independent at test time, which means that they can generalize over the robot workspace. Data collection and algorithm evaluation were done on a real robotic system.
\end{itemize}

\section{Related Work}
\label{sec:RelatedWork}
One of the main challenges in contact-rich assembly tasks, such as peg insertion, stacking, and edge following, is that the operation should be robust to the uncertainty in the target location. The reason is that even a slight misalignment would create very large contact forces that might cause the task to fail or even damages the parts or the robot itself.
Contact-rich assembly tasks, especially peg insertion, have been studied for a long time, due to their relevance in manufacturing. Previous work includes RL-based methods, such as \cite{lee2019making}, \cite{luo2018deep}, \cite{schoettler2019deep}, \cite{DBLP:journals/corr/LevineWA15,DBLP:journals/corr/abs-1708-04033,DBLP:journals/corr/abs-1906-05841}, LfD-based methods \cite{zhang2018deep}, \cite{ehlers2019imitating}, and combination of RL and LfD \cite{vecerik2019practical,rajeswaran2017learning,zhu2018reinforcement}. These previous papers use different combinations of sensing modalities. For instance, \cite{luo2018deep} studied the insertion of a rigid peg into a deformable hole, where the position of the hole relative to the robot is known, possibly with minor uncertainty. The authors combine information of the robot's state with filtered force/torque readings to form inputs to the controller. The controller trained in \cite{zhang2018deep} uses RGB-D images as input, while position information was used in an auxiliary loss function during training. Furthermore, \cite{vecerik2019practical} proposed a DDPG-based approach. The authors used positions, velocities, torques, and visual features as observation input, and compared fixed, not completely fixed, and fully randomized goal locations. It was shown that the agent can solve the fixed position task without visual features, but such features were necessary for solving the task with randomized socket position. In 
\cite{lee2019making}, a study on the importance of three sensor modalities --- RGB images, force readings, and proprioception --- for peg insertion tasks was conducted in simulation.

LfD is a widely used technique in the machine learning community, usually involving a human (expert) as a demonstrator to provide demonstration about how to perform a task. For example, \cite{zhang2018deep} applied Virtual Reality teleoperation to collect demonstrations, and behavior cloning was used to train visuomotor policies; \cite{DBLP:journals/corr/HesterVPLSPSDOA17} proposed deep Q-learning from demonstrations, which leverages small amounts of demonstration data to accelerate learning. The demonstration data is collected by a human and a controller. 
Authors in \cite{vecerik2019practical} gathered demonstrations by using a $6$D mouse (SpaceNavigator) and interpreted the motion as Cartesian commands, which they translated to joint actions using the known kinematics of the arm. 
In some other related work, human demonstrators directly guide the robot to finish the task by holding it. For example, \cite{Kober2011} starts with a few correct human demonstrations obtained by a human holding the robot arm to perform the task, and then letting the robot learn a policy through trial and error. In \cite{Grollman2011-ID819}, learning is initialized with two incorrect demonstrations by a human holding the robot. Learning proceeds through guided exploration around the demonstrations. Even though most existing methods use a human as a demonstrator, because industrial manipulators, like the one used in this paper, are not collaborative, and also due to the simplicity of obtaining demonstrations for this task, we make use of an automated controller to collect data. 

\section{Learning Methods}
\label{methods}
In this section, we discuss the neural network architectures we use for training the multi-modal policies, and then describe the behavioral cloning method used in our experiments.

\subsection{Neural Network-Based Controllers}
\label{networks}

We consider three different architectures of neural networks that all take as input raw observations, and provide as outputs the robot command actions. In particular, at each time step $t$, a network takes as input a subset of the following observation vector $o_{t} = (I_{t}, FT_{t}, ee_{t})$ where $I_{t} \in \mathbb{R}^{60\times60\times3}$ is an RGB image from a general purpose web camera mounted on the wrist of the robot (first-person view), $FT_{t}\in \mathbb{R}^{6}$ is a force-torque sensor measurement from a wrist-mounted sensor, and $ ee_{t}\in \mathbb{R}^{3}$ is the Cartesian position of the end-effector, which are proprioceptive measurements from the joint encoders built-in the robot arm. The outputs of all neural networks are the robot commands, defined as the desired change in position $ \hat{\Delta}_{t}\in \mathbb{R}^{3}$ in robot base coordinate frame. The three types of neural networks considered are described in the following.


\subsubsection{ResNet}\label{subsubsec:resnet} 
\begin{figure}[h]
\centering
\includegraphics[scale=0.1]{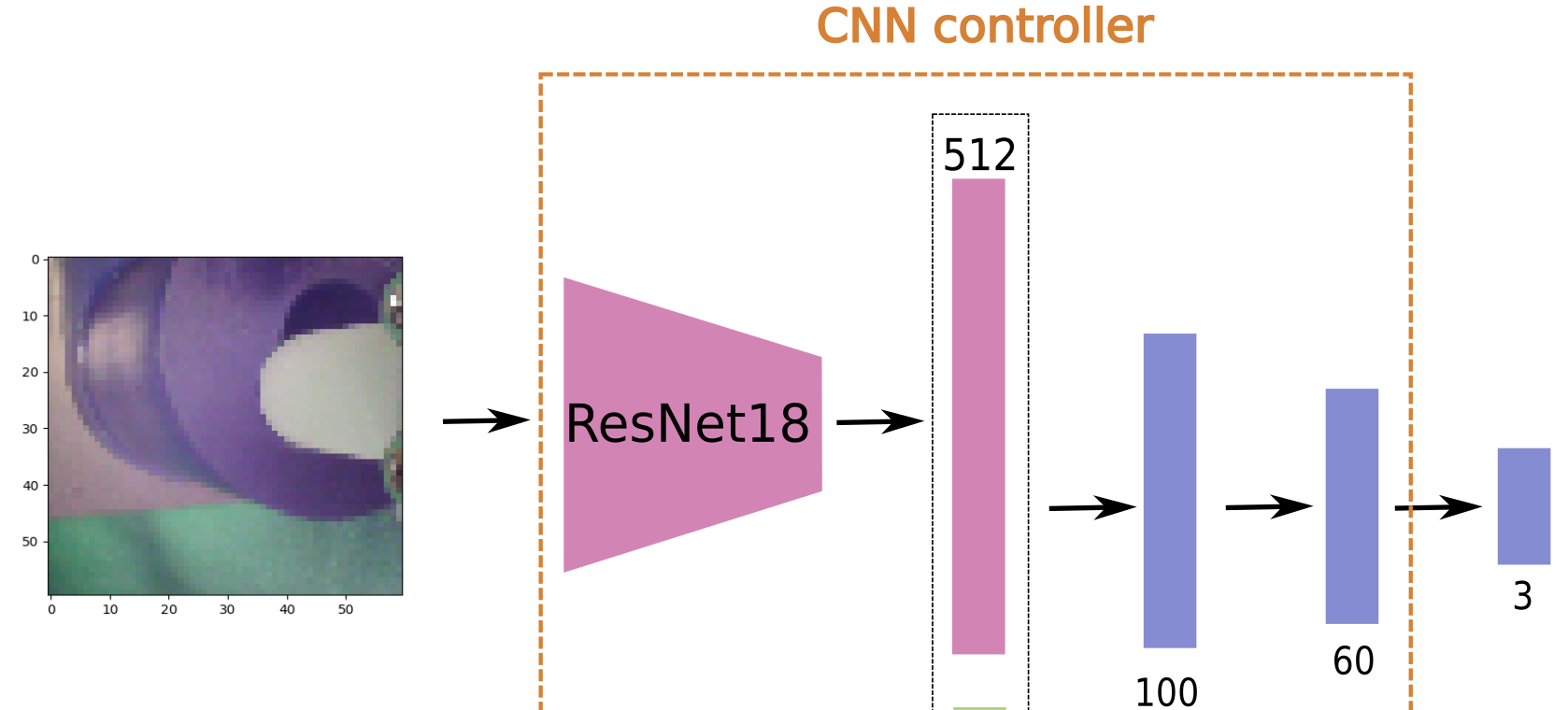}
\caption{Architecture of the ``ResNet" controller.} 
\label{fig:cnn3}
\end{figure}

The neural network is illustrated in Fig.~\ref{fig:cnn3}.
We use a pre-trained ResNet18 \cite{he2016deep} to extract a feature vector $ f^I_{t}\in \mathbb{R}^{512}$ from the image input $I_{t}$.
The features $ f^I_{t}$ are inputted into a network with 2 fully-connected layers that outputs the action vector $\Delta_t$, see Fig.~\ref{fig:cnn3}. 

\subsubsection{ResNet + raw sensory data}\label{subsubsec:resnet_raw} 
\begin{figure}[h]
\centering
\includegraphics[scale=0.1]{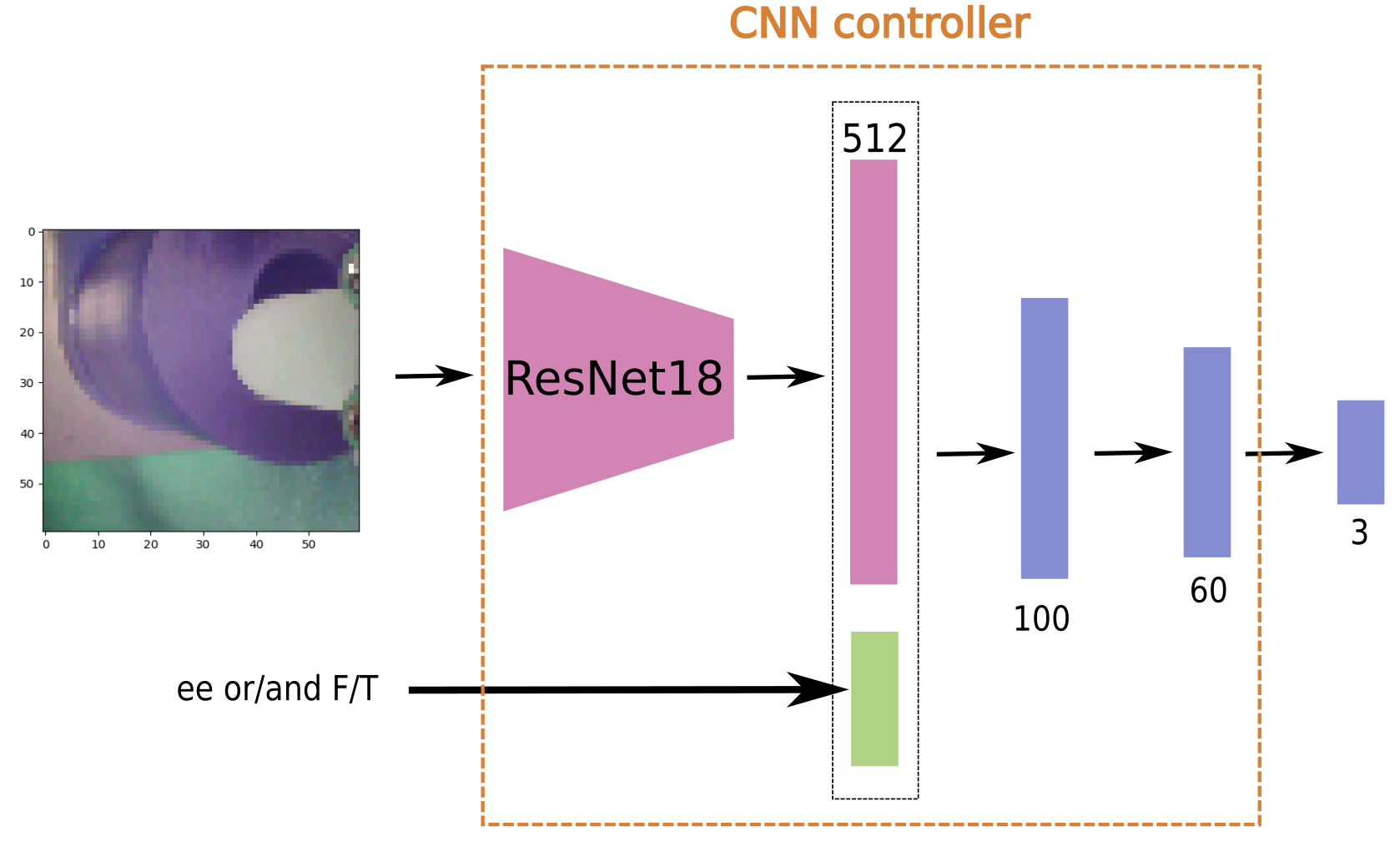}
\caption{Architecture of ``ResNet + raw sensory data" controller.} 
\label{fig:cnn2}
\end{figure}

The neural network is illustrated in Fig.~\ref{fig:cnn2}.
The same pre-trained ResNet18 \cite{he2016deep} is used to extract the feature vector $ f^I_{t}\in \mathbb{R}^{512}$ from the image input $I_{t}$.
The $FT_{t}$ and $ee_{t}$ raw sensory data are concatenated with $f^I_{t}$ into the feature vector $f^{NN}_{t}$, according to one of the following options that gives its name to several different control policies:
\begin{itemize}
    \item \textbf{ResNet + ee raw.} Consider image $I_{t}$ and $ee_{t}$ as input. Then $f^{NN}_{t}\in \mathbb{R}^{515}$.
    \item \textbf{ResNet + FT raw.} Consider image $I_{t}$ and $FT_{t}$ as input. Then $f^{NN}_{t}\in \mathbb{R}^{518}$.
    \item \textbf{ResNet + ee + FT raw.} Consider  image $I_{t}$, $ee_{t}$ and $FT_{t}$ as inputs. Then $f^{NN}_{t}\in \mathbb{R}^{521}$.
\end{itemize}{}
The feature vector $ f^{NN}_{t}$ is transformed into the action vector $\hat{\Delta}_t$ through a network with 2 fully-connected layers, see Fig.~\ref{fig:cnn2}. 

\subsubsection{ResNet + features extraction}\label{subsubsec:resnet_features} 
\begin{figure}[h]
\centering
\includegraphics[scale=0.1]{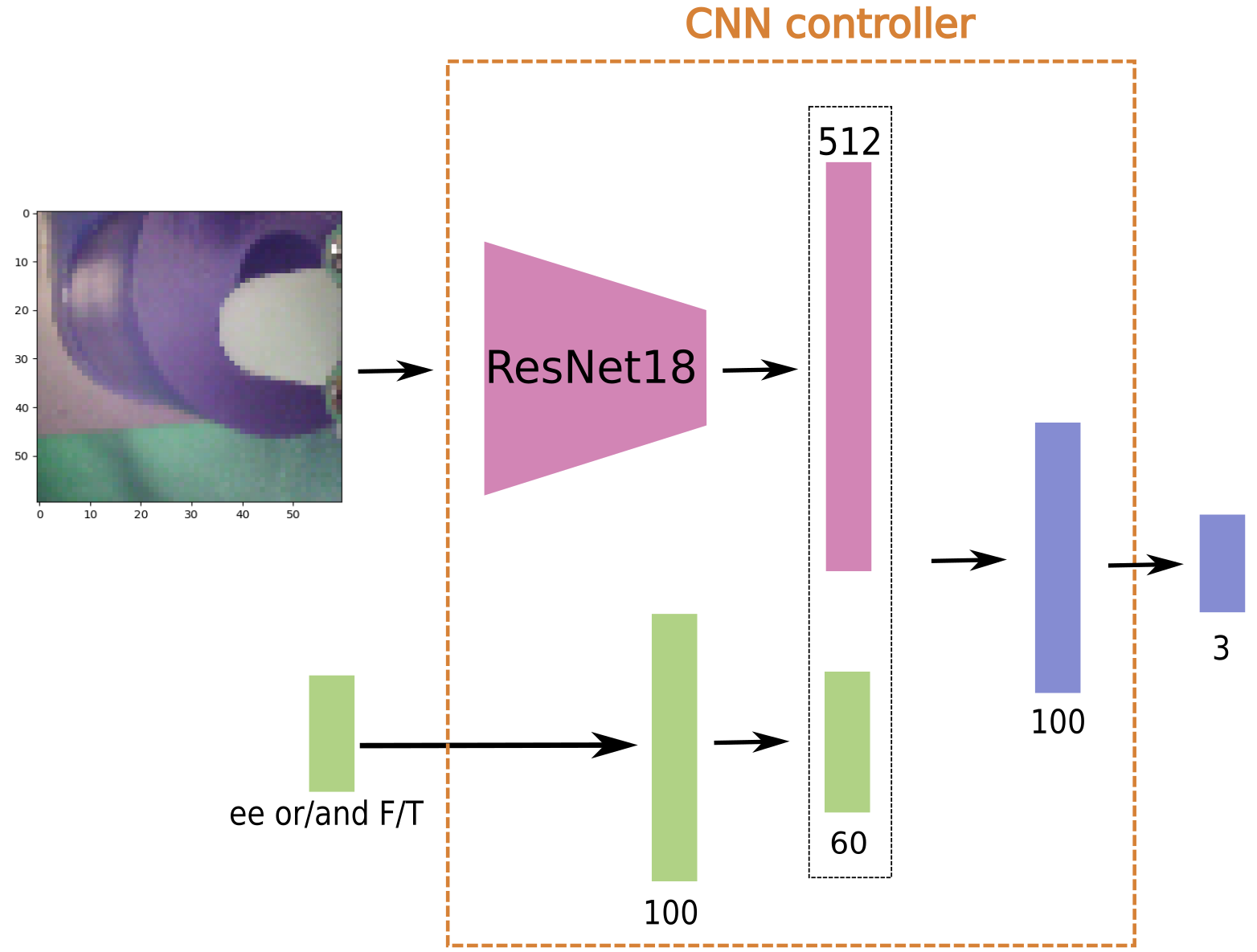}
\caption{Architecture of ``ResNet+features extraction" controller} 
\label{fig:cnn1}
\end{figure}

The neural network is illustrated in Fig.~\ref{fig:cnn1}.
The same pre-trained ResNet18 \cite{he2016deep} is used to extract the feature vector $ f^I_{t}\in \mathbb{R}^{512}$ from the image input $I_{t}$.
The $FT_{t}$ and $ee_{t}$ raw sensory data are inputted into a fully connected network, which outputs a feature vector $ f^r_{t}\in \mathbb{R}^{60}$, obtained by one of the following options that gives name to different control policies:
\begin{itemize}
    \item \textbf{ResNet + ee features.} Input to the fully connected network is $ee_{t}$.
    \item \textbf{ResNet + FT features.} Input to the fully connected network is $FT_{t}$.
    \item \textbf{ResNet + ee + FT features.} Input to the fully connected network is the concatenation of $ee_{t}$ and $FT_{t}$.
\end{itemize}{}
The feature vectors $ f^I_{t}$ and $ f^r_{t}$ are concatenated into the feature vector $ f^{NN}_{t}\in \mathbb{R}^{572}$, and $f^{NN}_{t}$ is transformed into the action vector $\hat{\Delta}_t$ with a fully connected network, see Fig.~\ref{fig:cnn1}.

The orange dash-line boxes depicted in each of the Fig.~\ref{fig:cnn3},\ref{fig:cnn2},\ref{fig:cnn1} are denoted as the ``CNN controller".

\subsubsection{Other controllers}\label{subsubsec:ee_features}
 Controllers without the vision input might also be considered. All the combinations $ee_t$, $FT_t$ and $ee_t \, +\, FT_t$ can be considered as inputs of a regression models or a neural network controller. For example we consider:

\begin{figure}[h]
\centering
\includegraphics[scale=0.3]{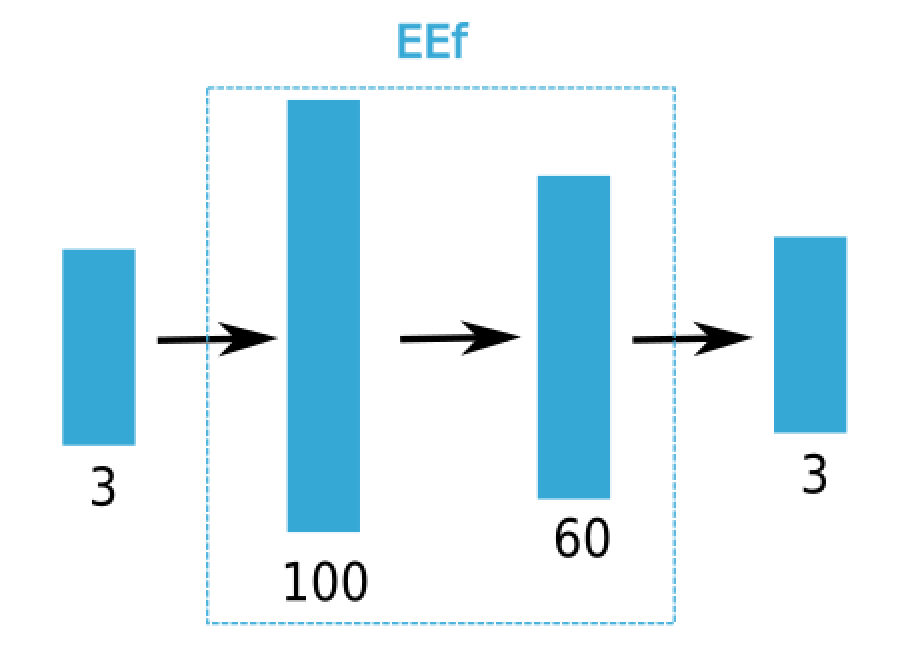}
\caption{Architecture of ``EEf" controller.} 
\label{fig:stepNN}
\end{figure}

\textbf{EE features.} The neural network is illustrated in Fig.~\ref{fig:stepNN}.

\begin{figure*}[t]
\centering
\includegraphics[scale=0.35]{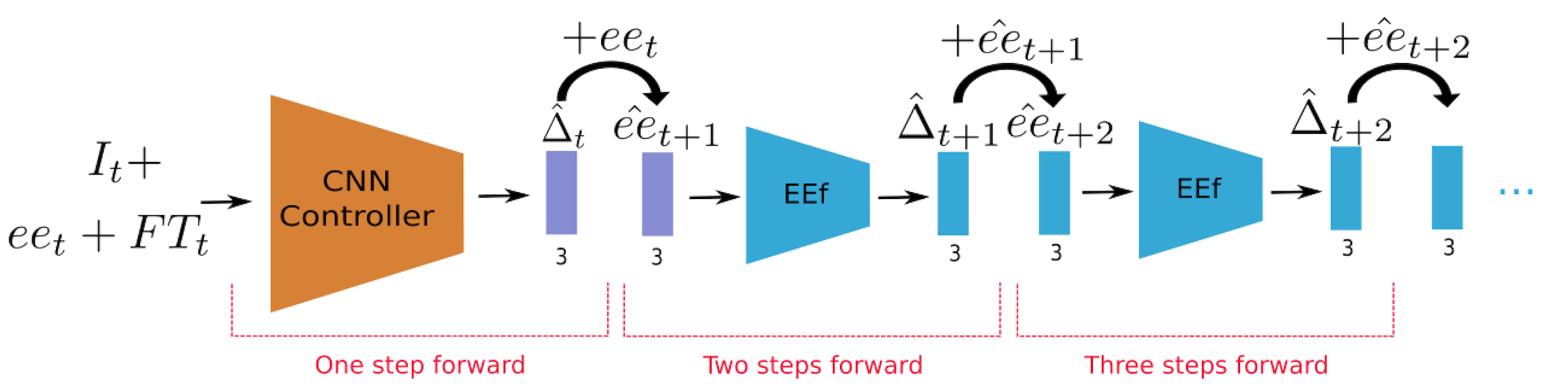}
\caption{Multi-step-ahead model. It shows the example with three steps. The ``CNN Controller" is used to predict the first step ahead, then the second and third steps are predicted by the ``EEf" model.} 
\label{fig:multi-step}
\end{figure*}

The $ee_{t}$ raw sensory data are inputted into a 2-layers fully connected network that outputs the action vector $\hat{\Delta}_t$. The blue dash-lined box in Fig.~\ref{fig:stepNN} is the network controller denoted by ``EEf".

\subsection{Multi-step-ahead Behavioral Cloning}
\label{behavior_cloning}

Behavioral cloning is a supervised learning method for imitation learning and decision making. Using behavioral cloning, the robot might learn satisfactory control policies without knowing either the system dynamics or the cost function. However, it sometimes suffers from a state distribution shift issue. The Data Aggregation (DAgger) algorithm can improve performance, but it requires a continual presence of the demonstrators to respond to queries each time the robot encounters a new state \cite{7487167}. 
To tackle the distribution mismatch problem without involving human supervisors, we propose to consider a multi-step-ahead loss function. The standard loss function used in behavioral cloning is the MSE of the action vector, used to match the demonstration of the expert as accurately as possible:
\begin{equation}\label{eq:mse}
    l = \frac{1}{T}\sum_{t=1}^T\Vert \hat{\Delta}_t - \Delta_t \Vert_2^2
\end{equation}{}
where $T$ is the number of training data examples.

We noticed that this loss function was not working well, because if the loss function was low in all the training points except in a few points the total loss function was still low, averaging out the loss in the critical points. In practice, during execution, the policy was not able to control the robot in some critical points, causing the experiment to fail. In order to incentivize policies that are accurate enough to control the robotic agent, we propose a loss function that considers multiple steps ahead, in open-loop fashion:
\begin{equation}\label{eq:mse_msa}
    l = \frac{1}{T}\sum_{t=1}^T\frac{1}{K}\sum_{k=1}^K\Vert \hat{\Delta}_{t+k} - \Delta_{t+k} \Vert_2^2
\end{equation}{}
where $K$ is the horizon of the multi-step-ahead prediction. In this way, the trained controller tends to guide the peg to an attraction area, which is within the distribution induced by the expert's demonstration. 
Consecutive predictions $\hat{\Delta}_{t+k}$ are obtained, as illustrated in Fig.~\ref{fig:multi-step}. The 1-step-ahead prediction $\hat{\Delta}_{t+1}$ is obtained by one of the CNN controllers described in Sec.~\ref{networks}. However, in order to predict the following actions $[\hat{\Delta}_{t+2},\ldots,\hat{\Delta}_{t+K}]$, the CNN controller would require an estimate of the observations at the future time instants $[\hat{o}_{t+1},\ldots,\hat{o}_{t+K-1}]$, which is a hard task (and, as we will see in the experiments, not really necessary). Instead, we propose to use the ``EEf" controller introduced in Sec~\ref{networks}, Fig.\ref{fig:stepNN}. The prediction at $t+k$ steps ahead with $k>1$ is computed as
\begin{equation*}
    \hat{ee}_{t+k} = \hat{ee}_{t+k-1}+EEf(\hat{ee}_{t+k-1}) = \hat{ee}_{t+k-1}+\hat{\Delta}_t
\end{equation*}{}
The multi-step-ahead model is shown in Fig. \ref{fig:multi-step}.
The proposed loss function can also be used in other learning frameworks. More complicated loss functions than Eq. \eqref{eq:mse} have been taken in consideration, too, see e.g. \cite{zhang2018deep}. Note that the multi-step-ahead loss functions could be applied to those loss functions, too.

\section{Experiments}
\label{exp}
The aim of the experiments described below is to empirically evaluate the different controllers described in Section~\ref{methods} based on different sensor modalities, and also investigate the effectiveness of the multi-step-ahead loss function proposed in Section~\ref{behavior_cloning}. In particular, this section will answer the following questions: 
\begin{itemize}
    \item Does the proposed multi-step-ahead loss function improve the training process compared to the traditional MSE loss function? 
    \item Is it better to concatenate the image features with the raw end-effector and force/torque measurements, or with the features extracted from these raw measurements obtained as in ``ResNet + features extraction"?
    \item Is it always beneficial to combine vision, force/torque readings, and end-effector positions into one controller?
\end{itemize}
To answer these questions, intensive experiments have been conducted on a real system. It is worth noticing that there is no comparison with other models, as the main goal of this paper is to understand different sensor modalities and evaluate if multi-step-ahead loss model improves the performance.

\subsection{Experimental Setup}
\label{setup}
Fig. \ref{fig:setup} shows the experimental setup. A MELFA RV-4FL robotic arm is used in the insertion task. The robotic arm is equipped with the following sensors: a Mitsubishi Electric 1F-FS001-W200 force/torque sensor attached to the wrist of the robot, a general-purpose RGB web camera mounted on top of the end-effector, and a set of built-in joint encoders that measure the robot's state.
The experiments are conducted using custom 3D-printed components that include a cylindrical peg, a peg holder mounted on the wrist after the force/torque sensor, and a (blue cylindrical) box with a hole in the center. The clearance between the peg and the hole is at the sub-millimeter level, measured at 0.7mm. The hole is attached to a heavy steel base to increase the stability of the hole's location during insertion. The robot is operated in position-control mode. During testing, the hole location is unknown, but is always in the field of view of the wrist camera. 

\subsection{Data Collection}
\label{data-collection}

An expert teacher needs to provide demonstrations. 
In PiH tasks, it is relatively easy to ascertain the location of the hole during training, so collecting data from an automated controller implemented on the robot is relatively simple and labor-efficient. For these reasons, the data in our experiments were collected with a basic controller in the following way.


Initially, we fix the location of the hole and randomly sample the initial position of the peg from a square area of size $40mm\times 40mm$ around the center of the hole, using a constant height of $20mm$ above the rim of the hole. After that, similar to the controllers studied in \cite{lee2019making} and \cite{ehlers2019imitating}, the automated controller executes three phases: approach, alignment, and insertion. The ``approach" step is the downward movement until the tip of the peg touches the surface of the platform. The ``alignment" is the step of sliding the peg on the surface of the platform along a straight line towards the center of the hole. 
The ``insertion" step is the movement straight down; it starts after the peg is aligned with the hole. By applying the automated controller, we collected $1,000$ complete insertions composed of the 3 steps and in total tens of thousands of data points. Among these insertions, 60\% are used for training, 20\% for validation, and 20\% for testing. Total data collection time was around 4.5 hours.

\subsection{Results}
\label{results}
In this section, we report evaluation results on the performance of the multi-step-ahead loss function for behavioral cloning proposed in Sec.\ref{behavior_cloning}, as well as the performance of the policy with multi-modal perceptual representations described in Sec.~\ref{networks}. The 7 controllers described in Sec.~\ref{subsubsec:resnet}, \ref{subsubsec:resnet_raw}, and \ref{subsubsec:resnet_features} are trained using the multi-step-ahead loss function with horizon $K=\{1,2,3\}$, for a total of 21 networks. Each controller has been tested for 20 insertions on the real system, randomizing the initial conditions of both the peg position and the hole location, for a total of 420 experiments. Notice, that the hole location is not fixed anymore and varies in the work space of the robot. The controllers that can handle different hole locations are called \textit{goal independent}. The controllers mentioned in Sec.~\ref{subsubsec:ee_features} were also tested, but are outside the definition of first-person view, and thus cannot generalize to different hole base locations. Therefore, for space reasons, their results are not reported. Still, it is important to mention the performance of the ``EEf'' controller, because this network is used in the loss function \eqref{eq:mse_msa}. There was high adherence to the data, with low values of the loss function in both the training and testing data. However, the policy performed poorly in the real experiments, because of lack of information about force and torque; the force measurements were exceeding the limits allowed by the robot, forcing the experiments to end. Nevertheless, the prediction accuracy of this controller justifies the usage of this controller in defining the multi-step-ahead loss function.


\begin{table*}[ht]
\centering
  \begin{tabular}{ c | l | c | c | c | c |}
  \midrule
    Network ID & Model Description & Full Insertion \% & Partial Insertion \% & Touch the Platform \% & Goal Independent\\ 
    \midrule
    1.1 & 1-step-ahead ResNet & 30 & 0 & 70 & \textbf{yes} \\ \cdashline{1-6}
    1.2 & 1-step-ahead ResNet + ee + FT & 55 & 15 & 30 & no \\ 
    1.3 & 1-step-ahead ResNet + ee & 50 & 0 & 50 & no \\ 
    1.4  & 1-step-ahead ResNet + FT & 75 & 0 & 25 & \textbf{yes} \\ \cdashline{1-6}
    1.5  & 1-step-ahead ResNet + ee + FT features & 55 & 10 & 35 & no \\ 
    1.6  & 1-step-ahead ResNet + ee features & 45 & 20 & 35 & no \\ 
    1.7  & 1-step-ahead ResNet +  FT features & 70 & 0 & 30 & \textbf{yes} \\ \midrule
    2.1 & 2-step-ahead ResNet& 95 & 0 & 5 &\textbf{yes} \\ \cdashline{1-6}
    2.2 & 2-step-ahead ResNet + ee + FT & 100 & 0 & 0 & no \\ 
    2.3 & 2-step-ahead ResNet + ee & 75 & 5 & 20 & no \\ 
    \textbf{2.4}  & \textbf{2-step-ahead ResNet + FT} & \textbf{100} & \textbf{0} & \textbf{0} & \textbf{yes} \\ \cdashline{1-6}
    2.5  & 2-step-ahead ResNet + ee + FT features & 85 & 5 & 10 & no \\
    2.6  & 2-step-ahead ResNet + ee features & 80 & 0 & 20 & no \\ 
    2.7  & 2-step-ahead ResNet + FT features & 95 & 0 & 5 & \textbf{yes} \\ 
    \midrule
    3.1  & 3-step-ahead ResNet & 85 & 0 & 15 & \textbf{yes} \\ \cdashline{1-6}
     3.2 & 3-step-ahead ResNet + ee + FT & 100 & 0 & 0 & no \\
    3.3 & 3-step-ahead ResNet + ee & 100 & 0 & 0 & no \\ 
    3.4  & 3-step-ahead ResNet + FT & 80 & 0 & 20 & \textbf{yes} \\ \cdashline{1-6}
    3.5  & 3-step-ahead ResNet + ee + FT features & 85 & 5 & 10 & no \\
    3.6  & 3-step-ahead ResNet + ee features & 90 & 0 & 10 & no \\ 
    \textbf{3.7}  & \textbf{3-step-ahead ResNet + FT features} & \textbf{100} & \textbf{0} & \textbf{0} & \textbf{yes} \\ \midrule
  \end{tabular}  \caption{Results of multi-modal perception, where each row corresponds to 20 insertions with random initial conditions of the peg and the hole positions. The first column is the network ID; the second column is the network model; Third to fifth column are the percentage of successful insertion; the last column indicates if the neural network model is goal independent.  Full Insertion means the peg can be successfully inserted into the hole; Partial Insertion means the peg aligns with the hole, but the insertion fails; Touch the Platform represents the peg can touch the platform, but fails to align with the hole.}
  \label{tab:2}
\end{table*}

Table \ref{tab:2} reports the resulting success and failure percentages for all  insertions. The table is organized in three blocks separated by solid lines, where each block is named according to the time horizon of the loss function: network IDs 1.1-1.7 are for the 1-step-ahead models, denoted as block 1; network IDs 2.1-2.7 are for the 2-step-ahead models, denoted as block 2; and network IDs 3.1-3.6 are for the 3-step-ahead models, denoted as block 3. Each block is divided into 3 sub-blocks separated by dashed lines: the first sub-block is the ``ResNet'' controller defined in Sec.~\ref{subsubsec:resnet}, in the second sub-block there are the ``ResNet + raw sensory data'' controllers defined in Sec.~\ref{subsubsec:resnet_raw}, and in the third sub-block, there are the ``ResNet + features extraction'' controllers defined in Sec.~\ref{subsubsec:resnet_features}.

\subsubsection{Effect of using different sensor modalities} Table \ref{tab:2} shows that for the standard one-step-ahead loss function, in block 1, adding the end-effector and/or force/torque readings improves the performance w.r.t. using only vision, ``ResNet'' network ID 1.1. However, we can see that considering the end-effector information is not always useful, and the controllers that consider ``ee" measurements, network IDs 1.2,1.4,1.5,1.6, actually perform worse than the controllers that are based only on vision and force/torque,``ResNet + FT raw/features''. 
This is likely due to the mismatch of the end-effector positions between training data and the test insertions, known as distribution shift. The high information content coming from the end-effector positions comes at the price of lack of generality of the controller, which becomes tied to specific spatial locations. Another advantage of not using the end-effector locations is that the controllers are \textit{goal independent}, there is no need to specify the $x,y,z$ location of the target hole. Indeed, for controllers not based on the ``ee" measurements, we were moving the target hole freely into the whole robot workspace at test time, and performing the insertions starting from a neighborhood ($40mm\times 40mm \times 20m$) around the hole center. 

\subsubsection{Effect of the multi-step-ahead loss function}\label{subsubsec:}
\begin{figure}[t]
\centering
\includegraphics[scale=0.23]{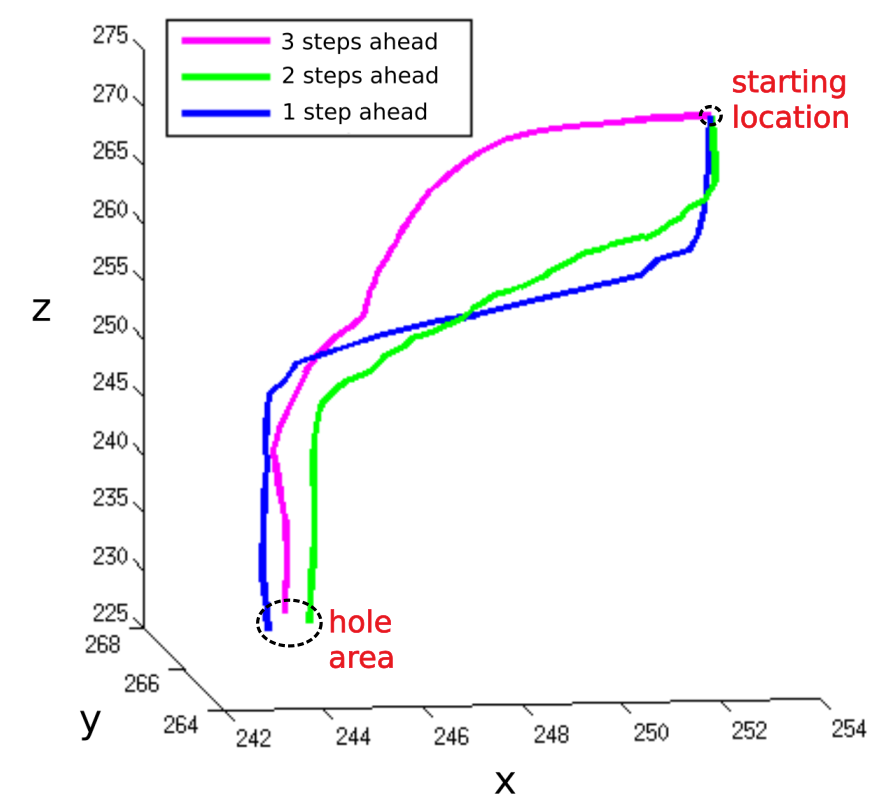}
\caption{Trajectories of different controllers. The blue, green and magenta lines are the 1-, 2-, and 3-step-ahead models, respectively. As indicated in the figure, the starting positions are at the top right corner, and the ending positions are at the bottom left.} 
\label{fig:traj}
\end{figure}

The idea behind the multi-step-ahead loss function is based on the findings described above. We want to exploit the high accuracy of the end-effector information during training, but avoid the high sensitivity and locality of this information at test time, which might worsen performance and limit the ability to generalize to different goal locations. 
By comparing all the networks in the three blocks, we find that 2-step-ahead models perform much better than the models trained by the original 1-step-ahead loss function in all the cases. Moreover, in general, the 3-step-ahead models work better than the models trained by the 2-step-ahead loss, except for network IDs 2.4 and 3.4. The ``ResNet'' controllers (network IDs 1.1, 2.1 and 3.1) show that vision inputs alone are insufficient to solve the task with standard 1-step-ahead loss function, having only 0.30\% success rate, but using the proposed multi-step-ahead function greatly improves performance. 
%
An example of insertion trajectories for the ``ResNet'' controller with loss function horizons $K=\{1,2,3\}$ is shown in Fig.\ref{fig:traj}. 

\subsubsection{Effect of using raw sensory data and features} In each of the three blocks, the ``ResNet + raw sensory data'' controllers tend to outperform the ``ResNet + features'' controllers independently of which sensory modalities were considered. However, the ``ResNet + features'' controllers have a monotonic increase in performance as the horizon of the multi-step-ahead loss function increases (compare network ID 1.5 with 2.5 and 3.5; ID 1.6 with 2.6 and 3.6; ID 1.7
with 2.7 and 3.7). This suggests that the lower performance of these controllers in blocks 1 and 2 might be related to the need for carefully training the extra layers in the networks, which is better obtained with the longest horizon of the multi-step-ahead loss function.


According to these results, the best controllers are the ``3-step-ahead ResNet + FT features'' and ``2-step-ahead ResNet + FT features'', achieving 100\% successful insertions in the whole workspace of the robot --- they both are goal-independent controllers.

\section{Conclusion and Future Work}
\label{conclusion-future}
We have performed and analyzed ablation studies on multi-modal perception for peg insertion tasks on a real robotic system  using a behavioral cloning approach. Some of the main conclusions we have drawn include: it is not always beneficial to combine vision, force/torque readings, and end-effector measurementd into one controller, due to the mismatch of the end-effector positions between training and testing the best performance are obtained without end-effector position and this grant goal independent controllers; it can be beneficial to map the raw force/torque and end-effector measurements to a well trained features space before using them for control; and, the proposed training method for behavioral cloning based on a multi-step-ahead loss function significantly improves the success rate of behavioral cloning. 

This paper shows our initial investigations related to PiH insertion tasks validated over 400 insertions at test time on a robotic system. However, further research could be done to validate and extend our findings on multi-modal perception on real robotic systems, and further generalize the proposed multi-step-ahead loss function. Possible future work includes, but is not limited to, the following directions:
\begin{itemize}
\item Extend the ablation studies for multi-perception to several other assembly tasks and investigate the generalization of the proposed multi-step-ahead loss function to other learning frameworks.
\item Add other sensor modalities: depth information is also a fundamental sensor input and widely used in assembly tasks \cite{zhang2018deep}; the vision input from a third person p.o.v.; touch sensors on the gripper fingers. 
\item The method could be improved by incorporating RL, possibly using the already proposed controllers as a warm-up policy and exploring the state space to improve performance.
\item It is still unclear how sensitive the drawn conclusions are w.r.t. variations in the architectures of the networks considered. In the future, we plan to add variations in our comparison framework, e.g. using different image processing network models, changing the number of fully connected layers, using different architectures, etc. 

\end{itemize}


\bibliographystyle{ieeetr}
\bibliography{icra2020_LfD.bib}  

\end{document}